\documentclass[runningheads]{llncs}

\usepackage{times}
\usepackage{epsfig}
\usepackage{graphicx}
\usepackage{amsmath}
\usepackage{amssymb}
\usepackage{color}
\usepackage[table]{xcolor}
\usepackage[lofdepth,lotdepth, caption=false,font=footnotesize]{subfig}
\usepackage{hhline}
\usepackage{cite}
\usepackage{stfloats}
\definecolor{Gray}{gray}{0.85}

\newcommand{\physics}{
\begin{figure}[t]
    \centering
    \includegraphics[width=0.8\linewidth]{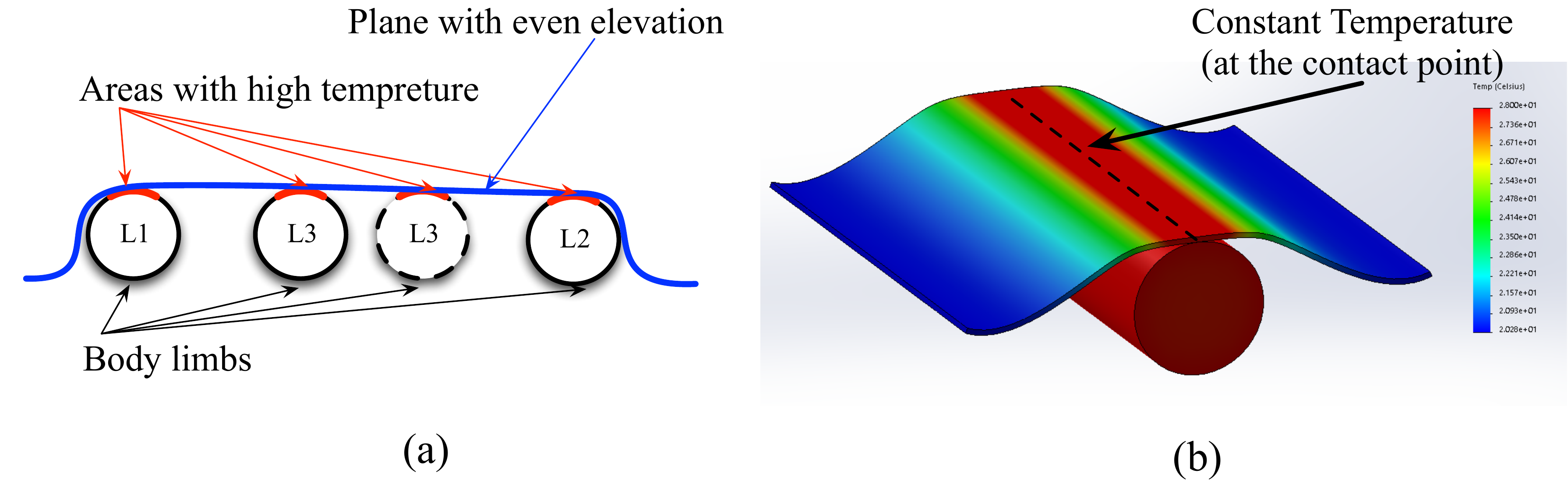}
    \caption{(a) Multiple limbs under a cover, (b) Temperature simulation of a covered cylinder with constant temperature.}
    \label{fig:physics}
    \vspace{-.2in}
\end{figure}
}

\newcommand{\multimodal}{
\begin{figure}[t]
    \centering
    \includegraphics[width=0.75\linewidth]{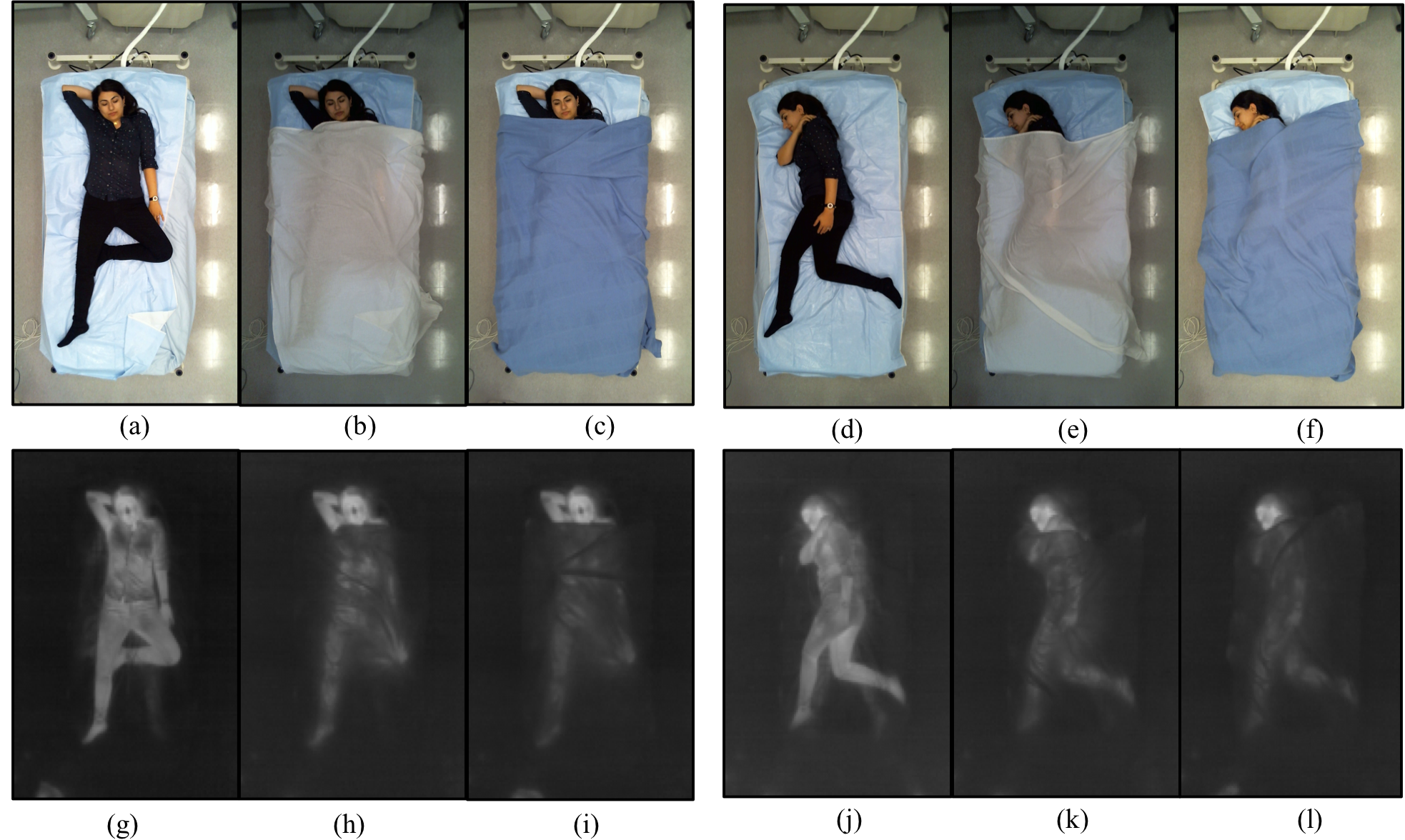}
    \caption{Samples from in-bed supine and side postures: (a-f) show images captured using RGB webcam, and (g-l) show images captured using LWIR camera. These images are taken from subject without cover and with two different types (one thin and one thick) of covers. }
    \label{fig:multimodal}
    \vspace{-.2in}
\end{figure}
}

\newcommand{\PCKrst}{
\begin{figure}[t]
    \centering
    \subfloat[]{\label{fig:danaPCK}\includegraphics[width=0.5\linewidth]{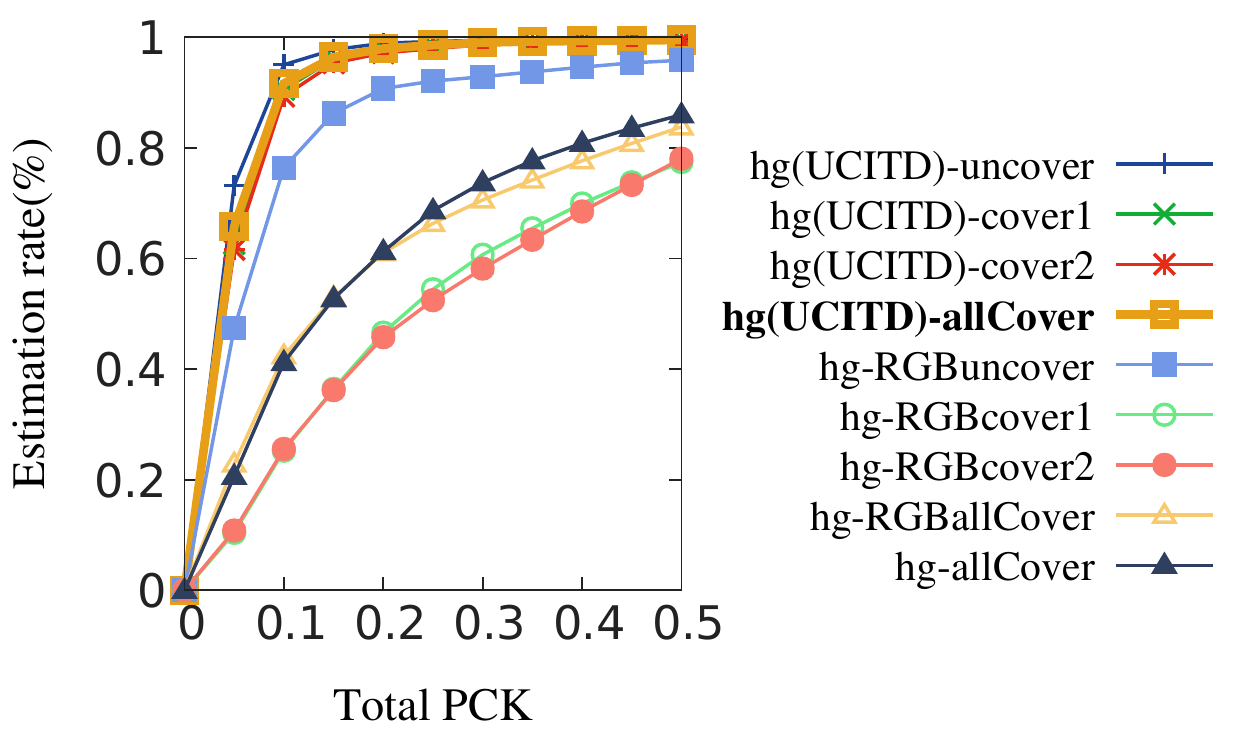}}
    \subfloat[]{\label{fig:simPCK}\includegraphics[width=0.5\linewidth]{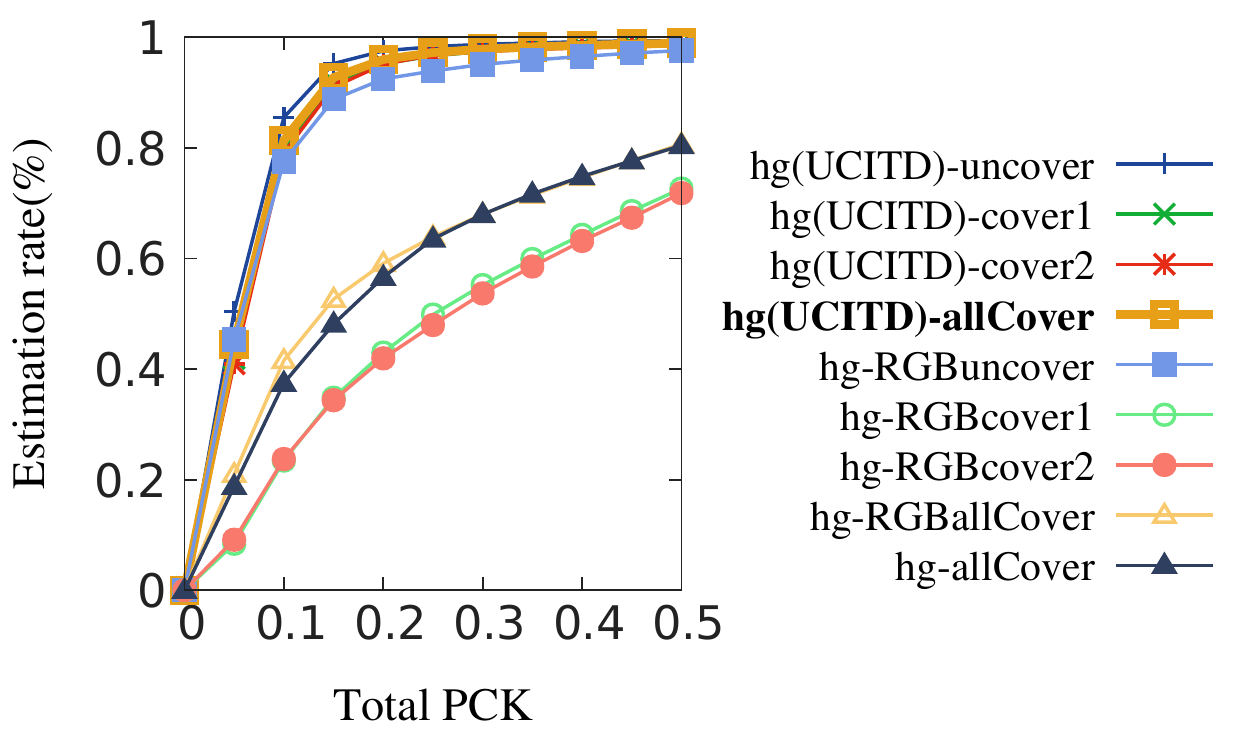}}
    \caption{PCK evaluation of in-bed human pose estimation models tested on data from (a) living room setting, and (b) hospital room setting, with different cover conditions. 
    For test samples, we use RGB to specify the RGB modality otherwise it will be the LWIR modality. 
    hg(UCITD) stands for the hourglass model trained via UCITD on SLP-LWIR set followed by different cover conditions.
    hg-allCover stands for applying pre-trained stacked hourglass (hg) model directly on SLP-LWIR test set with all cover conditions. hg-RGBallCover stands for applying pre-trained hg model directly on our SLP-RGB test set with all cover conditions.}
        \vspace{-.15in}
    \label{fig:PCKrst}
    \vspace{-.15in}
\end{figure}
}

\newcommand{\comparison}{
\begin{figure}[b]
 \centering
 \subfloat{\includegraphics[width=0.75\linewidth,{trim=0in 0in 0in 0in,
  clip=true}]{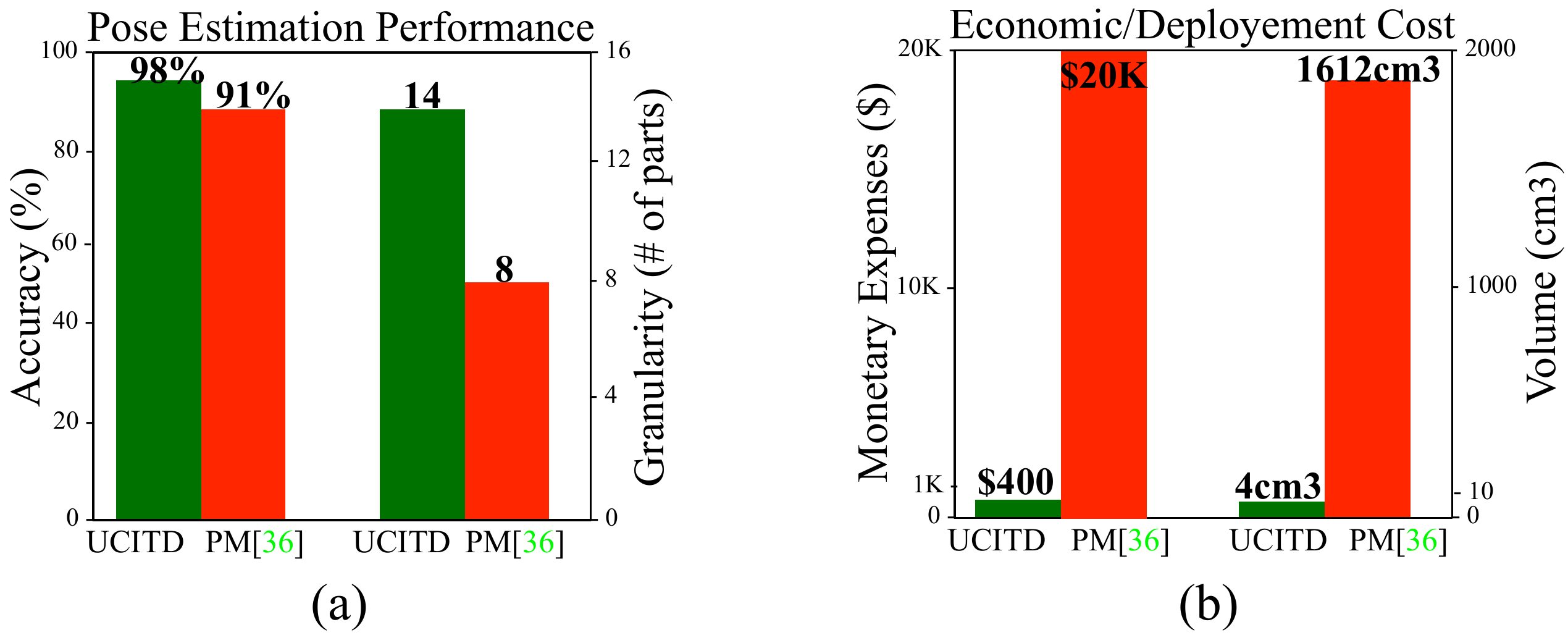}}
   \caption{Comparison between our UCITD and the PM-based pose estimation model presented in \cite{Ostadabbas2014}: (a) Pose estimation performance  in terms of pose detection accuracy and granularity, (b) Economic/deployment cost based on the monitory expense and the volume of each system.}
   \vspace{-.2in}
\label{fig:comparison}
\end{figure}
}

\newcommand{\figref}[1]{Fig.~\ref{fig:#1}}

\newcommand{\lemref}[1]{Lemma~\ref{lem:#1}}

\usepackage[pagebackref=true,breaklinks=true,letterpaper=true,colorlinks,bookmarks=false]{hyperref}

\begin{document}

\title{Seeing Under the Cover: A Physics Guided Learning Approach for In-Bed Pose Estimation\thanks{Supported by the NSF Award \#1755695. Source code and SLP dataset available at: \url{https://web.northeastern.edu/ostadabbas/2019/06/27/multimodal-in-bed-pose-estimation/}}}

%
%
\author{Shuangjun Liu\orcidID{0000-0002-2717-5789} 
\and
Sarah Ostadabbas\orcidID{0000-0002-2216-9988}} 
\authorrunning{S. Liu and S. Ostadabbas}
%
\institute{Augmented Cognition Lab (ACLab), Northeastern University, Boston, USA \\
\email{\{shuliu,ostadabbas\}@ece.neu.edu}\\
\url{https://web.northeastern.edu/ostadabbas/}}

\maketitle

\begin{abstract}
Human in-bed pose estimation has huge practical values in medical and healthcare applications yet still mainly relies on expensive pressure mapping (PM) solutions. In this paper, we introduce our novel physics inspired vision-based approach that addresses the challenging issues associated with the in-bed pose estimation problem including monitoring a fully covered person in complete darkness. We reformulated this problem using our proposed Under the Cover Imaging via Thermal Diffusion (UCITD) method to capture the high resolution pose information of the body even when it is fully covered by using a long wavelength IR technique. We proposed a physical hyperparameter concept through which we achieved high quality groundtruth pose labels in different modalities. A fully annotated in-bed pose dataset called Simultaneously-collected multimodal Lying Pose (SLP) is also formed/released with the same order of magnitude as most existing large-scale human pose datasets to support complex models' training and evaluation. A network trained from scratch on it and tested on two diverse settings, one in a living room and the other in a hospital room  showed pose estimation performance of 98.0\% and 96.0\% in PCK0.2 standard, respectively. Moreover, in a multi-factor comparison with a state-of-the art in-bed pose monitoring solution based on PM, our solution showed significant superiority in all practical aspects by being 60 times cheaper, 300 times smaller, while having higher pose recognition granularity and accuracy.






\end{abstract}
\section{Introduction}
The poses that we take while sleeping carry important information about our physical and mental health evident in growing research in the sleep monitoring field. These studies reveal that lying poses affect the symptoms of many complications such as sleep apnea \cite{lee2015changes}, pressure ulcers \cite{ostadabbas2011posture}, and even carpal tunnel syndrome \cite{mccabe2011preferred}. Moreover, patients in hospitals are usually required to maintain specific poses after certain surgeries to get a better recovery result. Therefor, long-term monitoring and automatically detecting in-bed poses are of critical interest in healthcare \cite{ostadabbas2012resource}. 

%
%
%
Currently, besides self-reports obtained from the patients and/or visual inspection by the caregivers, in-bed pose estimation methods mainly rely on the use of pressure mapping (PM) systems. 
Although PM-based methods are effective at localizing areas of increased pressure and even automatically classifying overall postures \cite{Ostadabbas2014}, the pressure sensing mats are expensive and require frequent maintenance, which have prevented PM pose monitoring solutions from achieving large-scale popularity.

By contrast, camera-based  methods for human pose estimation show great advantages including their low cost and ease of maintenance, yet are hindered by the natural  sleeping conditions including being fully covered in full darkness. 
To employ computer vision for in-bed activity monitoring, some groups exclusively focus on detection of particularly sparse actions such as leaving or getting into a bed \cite{ding2009bed}. Depth modal is also extensively employed for this application \cite{martinez2013bam}, yet is limited to simple activity recognition or recognizing very few body parts such as head and torso. A patient motion capture (MoCap) system was proposed in \cite{achilles2016patient} for 3D human pose estimation, however their experimental setup was never verified in a real setting for covered cases. Near infrared (IR) modality has also been employed \cite{liu2019bed} for long-term monitoring in full darkness, however it does not address the covered cases. 
Additionally, in the area of human in-bed pose estimation, there is no publicly-available dataset to train complex recognition models with acceptable generalizability, neither to fairly evaluate their performance.

In this paper, in contrast to the common RGB- or depth-based pose estimation methods, we propose a novel in-bed pose estimation technique based on a physics inspired imaging approach, which can effectively preserve human pose information in the imaging process, in complete darkness and even when the person is fully covered under a blanket. Our contributions in this paper can be summarized as follows: (1) reformulating the imaging process and proposing a passive thermal imaging method called Under the Cover Imaging via Thermal Diffusion (UCITD) based on a long wavelength IR (LWIR) technology; (2) proposing a physical hyperparameter concept that leads to quality multimodal groundtruth pose label generation; (3) building/publicly releasing the \emph{first-ever} fully annotated in-bed human pose dataset, called Simultaneously-collected multimodal Lying Pose (SLP) (reads as Sleep dataset) under different cover conditions, with the size equivalent to the existing large-scale human pose datasets to facilitate complex models' training and evaluation; (4) training a state-of-the-art pose estimation model from scratch using our SLP dataset, which showed high estimation performance comparable to the recent successful RGB-based human pose estimation models; and (5) comparing with the existing methods with equivalent capabilities, our solution demonstrates higher pose estimation accuracy and granularity, with only a fraction of cost and size. 


\section{In-bed Pose Estimation}

\subsection{Problem Formulation}
The major challenges that hinder the use of computer vision techniques for the in-bed pose estimation problem are monitoring in full darkness and potential cover conditions. To discover a proper imaging process capable of addressing these challenges, we reformulated the imaging process as follows. Let's assume the majority of the physical entities in the world (e.g. human body) can be modeled as articulated rigid bodies by ignoring  their non-rigid deformation. The physical world composed of $N$ rigid bodies then can be described by a world state model \cite{liu2018inner}, such that $W_s = \{\boldsymbol{\alpha_i}, \boldsymbol{\beta_i}, \phi(i,j)| i,j \in N \}$, where $\alpha_i$ and $\beta_i$  stand for the appearance and the pose of rigid body $i$, and $\phi(i,j)$ stands for the relationship between rigid bodies $i$ and $j$. For example, a human has $N$ (depending on the granularity of the template that we choose) articulated
limbs in which each limb can be considered a rigid body and the joints between the limbs follow the biomechanical constraints of the body. 
Assuming light source $S$ as the source of illumination, image $I$ can then be modeled as a function $I = I(W_s, S)$. 
Based on these assumptions, we argue a necessary condition to recognized covered object using \lemref{recog}.
\begin{lemma}
\label{lem:recog}
A physical object is recognizable after being covered by another physical entity, only if image of the world state after the cover being applied  $W_{s \rightarrow c}$ is conditioned on the characteristics of the target object. or equivalently, $ I = I(W_{s \rightarrow c}, S| \alpha_t, \beta_t) \neq I(W_{s \rightarrow c}, S) $ 
where, $\alpha_t, \beta_t$ stand for the target object's appearance and pose term.
\end{lemma}

In RGB domain, covered body movement will not always cause the cover change in image or the change is unnoticeable, so we can assume $I(W_{s \rightarrow c}, S_{RGB}) = I(W_{s \rightarrow c}, S_{RGB}| \alpha_t, \beta_t)$. This means the resultant RGB image will be independent of the target object's shape and pose when fully covered. In comparison, the  depth image of covered object is actually partially conditioned on the target object's pose and satisfies \lemref{recog} and work under darkness, hence it shows promising results for under cover conditions \cite{achilles2016patient}. Yet, ambiguity  exists as shown in \figref{physics}(a), where human limbs are represented by cylinders. If the cover is tightly stretched over L1 and L2, the depth map will be independent of the exact location of L3. 

\physics

\subsection{Under the Cover Imaging via Thermal Diffusion (UCITD)}
Although most imaging process is based on reflected light radiation, classical physics proves that all physical objects have their own radiation which can be approximated by their blackbody characteristics as Planck's radiation law.  
This law demonstrates two important characteristics: (1) blackbody has a specific spectrum and intensity that depends only on the body's temperature, and (2) at specific wavelength, object with higher temperature emits stronger radiation. The Planck’s radiation law provides insights to solve our  specific in-bed pose estimation problem, in which even though there is a lack of illumination from limbs under the cover, there will always be temperature differences between human body and the surrounding environment. The temperature of the skin of a healthy human is around $33^\circ$C, while clothing reduces the surface temperature to about $28^\circ$C, when the ambient temperature is $20^\circ$C. 
Planck's law shows that in this range of temperature, the corresponding radiation energy concentrates in long wavelength infrared (LWIR) spectrum which is around 8-–15$\mu$m. 
%
Although blanket are not transparent to the LWIR radiation, in our specific context, the contact between cover and body parts while lying in bed introduces another physical phenomenon called heat transfer, which dramatically alters the temperature distribution around the contacted areas. This phenomenon can be described by the diffusion equation $ \nabla^2T  = \frac{1}{a}\frac{\partial T}{\partial t}$ 
, where $T=T(x,y,z,t)$ is the temperature as a function of coordinates $(x,y,z)$ and time $t$, $a$ is the thermal diffusivity, and $\nabla^2$ is a Laplacian operator.


Exact modeling of covered human body is beyond the scope of this paper, so we
simulate such contact by simplifying each human limb as a cylinder with diameter 50mm which is covered by a thin physical layer with a thickness of 2mm in Solidworks (see \figref{physics}(b)). 
To set boundary conditions, we assume the contact point of the cover will turn into a constant temperature similar to the human clothes temperature ($\approx 28^{\circ}C$) after sufficient time. Heat will diffuse into environment which has constant temperature around $20^{\circ}C$. 
%
Such simplified model reveals that the contact point of a cover has the peak temperature. Furthermore, when a limb is covered with a sheet or a blanket, the location of the contact point directly depends on the shape and the location of the limb. In other words, the heat map will highly depend on the $\alpha$ and $\beta$ of the covered limbs, which satisfy the condition proposed in \lemref{recog} and endorses the feasibility of  LWIR for under the cover human pose estimation: $I = I(W_{s \rightarrow c}, S_{LWIR}| \alpha_t, \beta_t) \neq I(W_{s \rightarrow c}, S_{LWIR}) $. 
Admittedly, real case is much more complicated than our simplified model. There could be multiple peaks in contacting area due to the wrinkles in the cover. Nearby limbs will also result in more complex temperature profile due to the overlapped heating effect. But the dependency of the heat map over the limb's $\alpha$ and $\beta$ will still hold. As we can see, human like profiles in the \figref{multimodal} via thermal imaging 
(second row) is well recognizable even when it is fully covered with a thick blanket. \figref{physics}(a) also shows the advantage of LWIR over depth as the heated area of $L3$ will depend on its location as long as it is contacted by the cover. 
We call this imaging approach that satisfies \lemref{recog} under the cover imaging via thermal diffusion (UCITD). 


\section{UCITD Groundtruth Labeling}
Although human profile under the cover is visible via UCITD, the pose details are not always clearly recognizable by only looking at the LWIR images. Human annotators are likely to assign wrong pose labels when labelling LWIR images, which introduces noisy labels challenge to this problem. To address this issue, 
we cast the imaging process as a function that maps the physical entity and its cover into the image plane as $I= I(\alpha_t, \beta_t, \alpha_c, \beta_c)$, where $\alpha_t$, $\beta_t$, $\alpha_c$ and $\beta_c$ stand for the target's and cover's appearance and pose, respectively. In this formulation, $I$ could be the result of any of the feasible imaging  modalities such as $I \in \{I_{RGB}, I_{Depth}, I_{LWIR}, \dots\}$. 


\multimodal

A labeling process then can be defined as a function $L$ that maps the $I$ back to the target pose state $\beta_t$, such that the estimation target pose is $\hat{\beta_t} = L\Big(I(\alpha_t, \beta_t, \alpha_c, \beta_c)\Big)$. 
Error $E(\hat{\beta_t}, \beta_t; \alpha_t, \alpha_c, \beta_c)$  depends on not only the pose terms but also the appearance terms.  As all these parameters (i.e. $\{ \alpha_t, \alpha_c, \beta_c\}$) can be decoupled from $\beta_t$ \cite{liu2018inner}, they can be deemed  as the hyperparameters of function $L$ and we can formulate the problems as an optimization: $\displaystyle{\min_{\alpha_t, \alpha_c, \beta_c} E}$. Unlike commonly referred hyperparameters in mathematical modeling, these variables are directly related to the physical properties of the object, so we call them \textit{physical hyperparameters}. 
Due to the physical constraints, we can not adjust them freely like other hyperparameters, yet we showed that in our application, physical hyperparameters can also be altered effectively to optimize target $L$ performance. 
Based on this formulation, we propose the following three guidelines to achieve a robust LWIR image groundtruth labeling. A labeling tool implemented following these guidelines will be released together with the paper. 


\noindent \textit{\underline{Guideline I:}} Perform labeling under settings with same $\beta_t$ but 
no cover to yield best pose labeling performance.

\noindent \textit{\underline{Guideline II:}} Employ  $I_{RGB}$ counterpart as a  guide to prune out false poses in $I_{LWIR}$.

\noindent \textit{\underline{Guideline III:}} When finding exact joint locations are intractable in one domain, employ labels from other domain with bounded bias via homography mapping.

\section{Under the Cover Pose Estimation Evaluation}
\label{sec:evaluation}

\subsection{SLP Dataset Description}
We setup two evaluation environments for our experiments, one in a lab setting turned into a regular living room and one in a simulated hospital room at Northeastern University Health Science Department. In each room, we mounted one RGB camera (a regular webcam) and one LWIR camera (a FLIR thermal imaging system) on the ceiling, where both were vertically aligned and adjacent to each other to keep small distance (detailed setup in the supplementary material). 
%
%
Using an IRB-approved protocol, we collected pose data from 102 subjects in the living room (called ``Room'' setting) and 7 volunteers in the hospital room (called ``Hosp'' setting), while lying in a bed and randomly changing their poses under three main categories of supine, left side, and right side. For each category, 15 poses are collected. For each pose, we altered the physical hyperparameters of the setting via manual intervention. We collected the images from both RGB and LWIR camera simultaneously to alter the function $I$. Moreover, we changed the cover condition from uncover, to cover one (a thin sheet with $\approx$1mm thickness), and then to cover two (a thick blanket with $\approx$3mm thickness) to alter $\alpha_c$ and $\beta_c$. In each cover condition, we waited around 10--20 seconds to mimic a stabilized pose during a real-life monitoring scenario. 
We follow pose definition of LSP \cite{johnson2010clustered} with 14 joints.
Data collection in the living room and hospital room allowed us to form our Simultaneously-collected multimodal Lying Pose (SLP) dataset which will be public released with the paper. SLP dataset is collected under two different settings with 13,770 samples for ``Room'' and 945 for ``Hosp'', among which first 90 subjects with 12150 samples of ``Room'' set are for training and rest 12 subjects with 1620 samples are for testing. ``Hosp'' set is used for test purpose only to show its field performance under  a simulated real application scenario.  
It is worth to mention that our SLP in-bed pose dataset has equivalent magnitude to most large-scale human pose dataset such as MPII \cite{andriluka20142d} and LSP \cite{johnson2010clustered}, which allows complex models' training from scratch. 

\subsection{In-bed Human Pose Estimation Performance}
To evaluate the pose estimation performance of the proposed pipeline, we trained a state-of-the-art 2D human pose estimation model from scratch, the stacked hourglass (hg) network \cite{newell2016stacked} which is one of top performance model for single human pose, using the LWIR images from 90 subjects training set 
in ``Room'' dataset with 8000 iterations, 30 epochs, and learning rate of 2.5e-4 as original settings for RGB domain. To investigate the effect of different cover conditions on model performance, we use probability of correct keypoint (PCK) as our metric which is extensively employed for human pose estimation \cite{johnson2010clustered, andriluka20142d}. In PCK metric, the distance between the estimated joint position and the ground-truth position is compared against a threshold defined as fraction of the person's torso length to form a joint level detection rate. 
Estimation results from each cover condition is reported.  
Due to the lack of public benchmark in this specific application, it is hard to provide a strict quantitative comparison. To launch a fair comparison, we fed pre-trained hg model with RGB samples without cover which suppose to have not only exact same uncover RGB domain as pre-trained model but also have similar ``pose hardness'' as test samples on UCITD. According to the evaluation shown in \figref{PCKrst}(a), our model demonstrated a 98.0\% accuracy at PCK0.2 which is marginally higher than pre-trained hg in RGB domain. Please note that this cross-domain comparison is between our UCITD under darkness and covered condition against a well-illuminated RGB condition for the other models. Test result also shows that when subjects are covered， tremendous impact is imposed on RGB hg but has only slight adverse effect on UCITD. Failure cases usually come when limbs are cuddled together that limbs will be misaligned to nearby body area due to the similar temperature of human profile or at transient moment caused by heat residue on bed.   

\PCKrst 

\subsection{Domain Adaptation Evaluation}


Many datasets for healthcare applications are collected in extremely controlled environments, which limit their learning transferability to the real-life settings due to the gap between simulated and real-world data distributions (i.e. domain shift).
With this consideration, to reveal the true performance of our technique in a practical application, we simulated a new deployment scenario by re-setting up the whole system in a hospital room, called ``Hosp'' setting. 
We intentionally altered all of the environment hyperparameters from the training set: (1) using a common hospital bed and mattress in ``Hosp'' setting, as supposed to a twin size metal bed-frame with middle firmness spring mattress used in ``Room'' setting; (2) repurchasing all covers to alter the cover appearance $\alpha_c$; (3) collecting data from new subjects to introduce a new $\alpha_t$; and (4) having different bed and room height which introduced a varied target distance from camera. We believe these are the most possibly changed parameters that can be seen in a real-world application. With a test dataset collected under the ``Hosp'' setting, our hg-trans model still showed 96.0\% pose estimation performance over PCK0.2 as shown in \figref{simPCK}, which is also marginally higher than its pre-trained RGB counterpart. 


\comparison
\subsection{Comparison with PM-based Pose Estimation}
Due to the lack of public benchmark and varying capabilities of different in-bed human pose estimation methods, we can hardly run an extensive comparison study. For a fair comparison, we believe the candidates should have following characteristics: (1) similar capabilities and pose granularity, and (2) concrete evaluation over real data instead of synthetic ones to reflect its practical value. Accordingly, candidate methods should be able to recognize multiple body parts simultaneously in darkness and varying cover conditions. As far as we know, none of the existing method can perfectly address all aforementioned standards except PM-based method in which the best one turns out to be the work by \cite{Ostadabbas2014} with comparable granularity. 
With close accuracy performance as shown in \figref{comparison}(a), our method shows higher granularity recognition ability with more joints being detected (14 vs. 8). Furthermore, authors in \cite{Ostadabbas2014} evaluated accuracy via visual inspection for overlapping area, while UCITD employs widely acknowledged PCK0.2 metric for human pose estimation study. 


Besides pose estimation performance, we further evaluated the cost-efficiency of UCITD against PM-based methods. We used a FLIR camera with 120$\times$160 resolution. For equivalent resolution, we employed Tekscan$^\circledR$ full body PM system with 192$\times$84 sensor resolution. The price and space cost comparison is shown in \figref{comparison}(b). Our UCITD approach achieves tremendous cost efficiency, by being 60 times cheaper (a rough cost estimation is around \$400) and 300 times smaller compared to the most advanced PM-based approach.
Furthermore, by using unidentifiable heat map, UCITD is both privacy-reserving and radiation-free. As a contact-less method it requires much less maintenance compared to the PM-base approaches, which are prone to failure due to pressure sensors drift over time. Due to the small form-factor of the UCITD technology, it can be mounted unobtrusively in any indoor environment to be used in long-term in-bed pose monitoring applications. 

It is worth mentioning that for other modalities (e.g. RGB), according to the \lemref{recog} if the covered image is conditioned on the underlying pose $\alpha_c$, it is possible to estimate the pose accurately. This point is apparent in the images in the first row of the \figref{multimodal}, in which the covered poses can  still be inferred from the RGB images that reveal the human profiles. However, one can imagine such condition will no longer hold for RGB modality in the full darkness and the need for another modality such as LWIR for in-bed pose monitoring is inevitable. 

\bibliographystyle{splncs04}
\bibliography{paper}

\end{document}